\documentclass[10pt,twocolumn,letterpaper]{article}

\usepackage{cvpr}
\usepackage{times}
\usepackage{epsfig}
\usepackage{amsmath}
\usepackage{amssymb}

\usepackage{url}
\usepackage{comment}
\usepackage{algorithm,algorithmic,graphicx}
\usepackage{bm}
\usepackage{booktabs}

\usepackage{caption}
\usepackage{subcaption}

\usepackage[pagebackref=true,breaklinks=true,letterpaper=true,colorlinks,bookmarks=false]{hyperref}

 \cvprfinalcopy 

\def\cvprPaperID{2265} 

\ifcvprfinal\fi

\begin{document}

\title{Stacked Attention Networks for Image Question Answering}

\author{
Zichao Yang$^{1}$,
Xiaodong He$^{2}$,
Jianfeng Gao$^{2}$,
Li Deng$^{2}$,
Alex Smola$^{1}$ \\
$^{1}$Carnegie Mellon University,
$^{2}$Microsoft Research, Redmond, WA 98052, USA\\
\texttt{zichaoy@cs.cmu.edu,
\{xiaohe, jfgao, deng\}@microsoft.com,
alex@smola.org}
}

\maketitle

\begin{abstract}
  This paper presents stacked attention networks (SANs) that learn to answer
  natural language questions from images. SANs use semantic representation of a
  question as query to search for the regions in an image that are related to
  the answer. We argue that image question answering (QA) often requires
  multiple steps of reasoning. Thus, we develop a multiple-layer SAN in which
  we query an image multiple times to infer the answer
  progressively. Experiments conducted on four image QA data sets demonstrate
  that the proposed SANs significantly outperform previous state-of-the-art
  approaches. The visualization of the attention layers illustrates the
  progress that the SAN locates the relevant visual clues that lead to the
  answer of the question layer-by-layer.
\end{abstract}

\vspace{-0.3cm}
\section{Introduction}
With the recent advancement in computer vision and in natural language
processing (NLP), image question answering (QA) becomes one of the most active
research areas~\cite{gao2015you, ren2015imageqa, malinowski2014multi,
  antol2015vqa, malinowski2015ask}. Unlike pure language based QA systems that
have been studied extensively in the NLP community~\cite{weston2014memory,
  kumar2015ask, bordes2014question, yih2015semantic, berant2014semantic,
  yih2014semantic}, image QA systems are designed to automatically answer
natural language questions according to the content of a reference image.

\begin{figure}[tbh]
  \centering
  \begin{subfigure}[t]{1.0\linewidth}
    \includegraphics[width=1.1\linewidth]{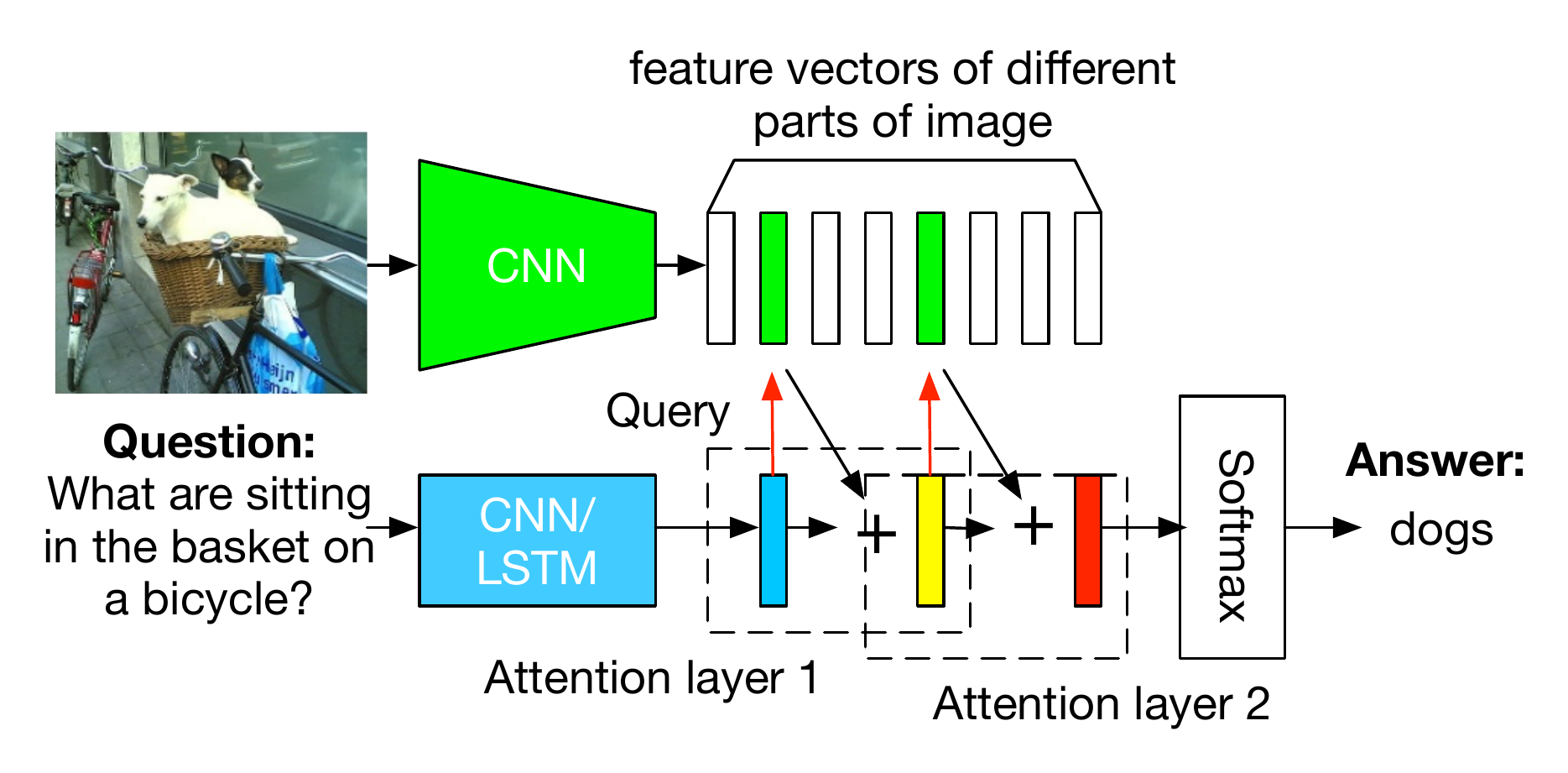}
    \centering
    \caption{Stacked Attention Network for Image QA}
    \label{fig:vqa_attention}
  \end{subfigure}

  \begin{subfigure}[t]{1.0\linewidth}
    \includegraphics[width=\linewidth]{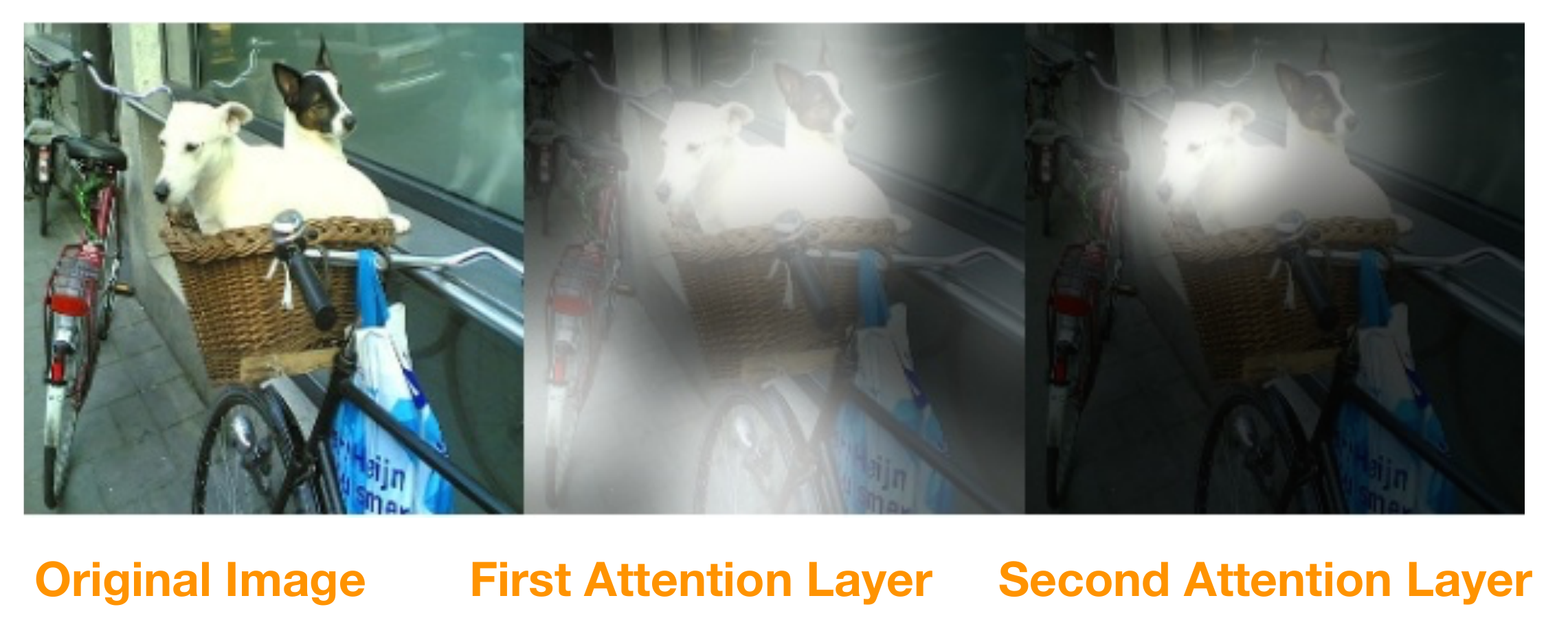}
    \centering
    \caption{Visualization of the learned multiple attention layers. The
      stacked attention network first focuses on all referred concepts, e.g.,
      \texttt{bicycle, basket} and objects in the basket (\texttt{dogs}) in the
      first attention layer and then further narrows down the focus in the
      second layer and finds out the answer \texttt{dog}.}
    \label{fig:example}
  \end{subfigure}
  \caption{Model architecture and visualization}
  \label{fig:model_example}
 \vspace{-0.5cm}
\end{figure}

Most of the recently proposed image QA models are based on neural networks
\cite{gao2015you, ren2015imageqa, malinowski2014multi, antol2015vqa,
  malinowski2015ask}. A commonly used approach was to extract a global image
feature vector using a convolution neural network
(CNN)~\cite{lecun1998gradient} and encode the corresponding question as a
feature vector using a long short-term memory network
(LSTM)~\cite{hochreiter1997long} and then combine them to infer the
answer. Though impressive results have been reported, these models often fail
to give precise answers when such answers are related to a set of
\emph{fine-grained} regions in an image.

By examining the image QA data sets, we find that it is often that case that
answering a question from an image requires multi-step reasoning. Take the
question and image in Fig.~\ref{fig:model_example} as an example. There are
several objects in the image: \texttt{bicycles, window, street, baskets} and
\texttt{dogs}. To answer the question \texttt{what are sitting in the basket on
  a bicycle}, we need to first locate those objects (e.g. \texttt{basket,
  bicycle}) and concepts (e.g., \texttt{sitting in}) referred in the question,
then gradually rule out irrelevant objects, and finally pinpoint to the region
that are most indicative to infer the answer (i.e., \texttt{dogs} in the
example).

In this paper, we propose stacked attention networks (SANs) that allow
multi-step reasoning for image QA. SANs can be viewed as an extension of the
attention mechanism that has been successfully applied in image
captioning~\cite{xu2015show} and machine
translation~\cite{bahdanau2014neural}. The overall architecture of SAN is
illustrated in Fig.~\ref{fig:vqa_attention}. The SAN consists of three major
components: (1) the image model, which uses a CNN to extract high level image
representations, e.g. one vector for each region of the image; (2) the question
model, which uses a CNN or a LSTM to extract a semantic vector of the question
and (3) the stacked attention model, which locates, via multi-step reasoning,
the image regions that are relevant to the question for answer prediction. As
illustrated in Fig.~\ref{fig:vqa_attention}, the SAN first uses the question
vector to query the image vectors in the first visual attention layer, then
combine the question vector and the retrieved image vectors to form a refined
query vector to query the image vectors again in the second attention
layer. The higher-level attention layer gives a sharper attention distribution
focusing on the regions that are more relevant to the answer. Finally, we
combine the image features from the highest attention layer with the last query
vector to predict the answer.

The main contributions of our work are three-fold. First, we propose a stacked
attention network for image QA tasks. Second, we perform comprehensive
evaluations on four image QA benchmarks, demonstrating that the proposed
multiple-layer SAN outperforms previous state-of-the-art approaches by a
substantial margin. Third, we perform a detailed analysis where we visualize
the outputs of different attention layers of the SAN and demonstrate the
process that the SAN takes multiple steps to progressively focus the attention
on the relevant visual clues that lead to the answer.

\vspace{-0.3cm}
\section{Related Work}
Image QA is closely related to image captioning~\cite{chen2014learning,
  xu2015show, fang2014captions, vinyals2014show, kiros2014unifying,
  karpathy2014deep, mao2014deep}. In \cite{vinyals2014show}, the system first
extracted a high level image feature vector from GoogleNet and then fed it into
a LSTM to generate captions. The method proposed in \cite{xu2015show} went one
step further to use an attention mechanism in the caption generation
process. Different from \cite{xu2015show, vinyals2014show}, the approach
proposed in \cite{fang2014captions} first used a CNN to detect words given the
images, then used a maximum entropy language model to generate a list of
caption candidates, and finally used a deep multimodal similarity model (DMSM)
to re-rank the candidates. Instead of using a RNN or a LSTM, the DMSM uses a
CNN to model the semantics of captions.

Unlike image captioning, in image QA, the question is given and the task is to
learn the relevant visual and text representation to infer the answer.  In
order to facilitate the research of image QA, several data sets have been
constructed in~\cite{malinowski2015ask, ren2015imageqa, gao2015you,
  antol2015vqa} either through automatic generation based on image caption data
or by human labeling of questions and answers given images. Among them, the
image QA data set in~\cite{ren2015imageqa} is generated based on the COCO
caption data set. Given a sentence that describes an image, the authors first
used a parser to parse the sentence, then replaced the key word in the sentence
using question words and the key word became the answer. \cite{gao2015you}
created an image QA data set through human labeling. The initial version was in
Chinese and then was translated to English. \cite{antol2015vqa} also created an
image QA data set through human labeling. They collected questions and answers
not only for real images, but also for abstract scenes.

Several image QA models were proposed in the
literature. \cite{malinowski2014multi} used semantic parsers and image
segmentation methods to predict answers based on images and
questions. \cite{malinowski2015ask, gao2015you} both used encoder-decoder
framework to generate answers given images and questions. They first used a
LSTM to encoder the images and questions and then used another LSTM to decode
the answers. They both fed the image feature to every LSTM
cell. \cite{ren2015imageqa} proposed several neural network based models,
including the encoder-decoder based models that use single direction LSTMs and
bi-direction LSTMs, respectively. However, the authors found the concatenation
of image features and bag of words features worked the best.
\cite{antol2015vqa} first encoded questions with LSTMs and then combined
question vectors with image vectors by element wise
multiplication. \cite{ma2015learning} used a CNN for question modeling and used
convolution operations to combine question vectors and image feature
vectors. We compare the SAN with these models in Sec.~\ref{sec:experiments}.

To the best of our knowledge, the attention mechanism, which has been proved
very successful in image captioning, has not been explored for image QA. The
SAN adapt the attention mechanism to image QA, and can be viewed as a
significant extension to previous models~\cite{xu2015show} in that multiple
attention layers are used to support multi-step reasoning for the image QA
task.

\vspace{-0.2cm}
\section{Stacked Attention Networks (SANs)}
\label{sec:model}

The overall architecture of the SAN is shown in
Fig.~\ref{fig:vqa_attention}. We describe the three major components of SAN in
this section: the image model, the question model, and the stacked attention
model.

\vspace{-0.2cm}
\subsection{Image Model}
The image model uses a CNN~\cite{krizhevsky2012imagenet, simonyan2014very,
  szegedy2014going} to get the representation of images. Specifically, the
VGGNet~\cite{simonyan2014very} is used to extract the image feature map $f_{I}$
from a raw image $I$:
\begin{figure}[tbh]
  \vspace{-0.5cm}
    \includegraphics[width=0.6\linewidth]{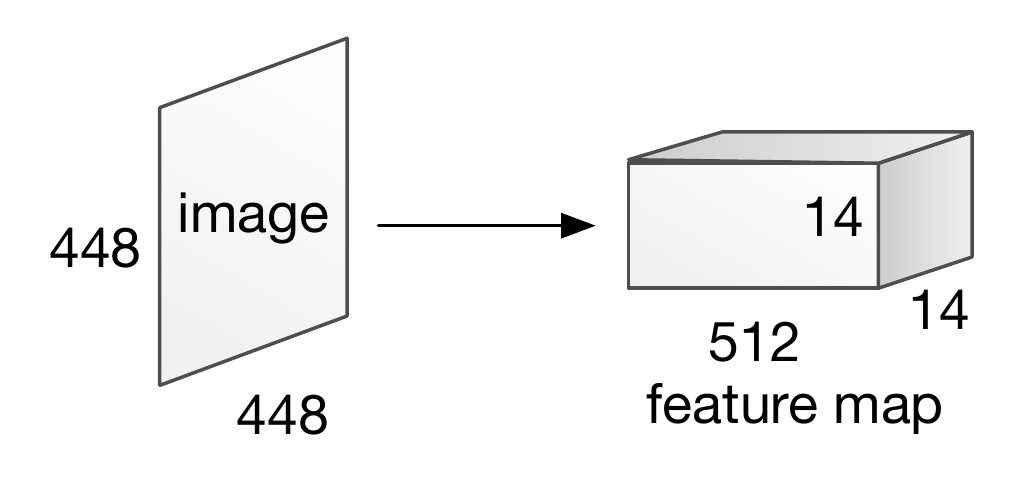}
    \centering
    \caption{CNN based image model}
    \label{fig:cnn_img}
  \vspace{-0.5cm}
\end{figure}

\begin{equation}
  f_{I} = \text{CNN}_{vgg}(I).
\end{equation}
Unlike previous studies~\cite{ren2015imageqa,ma2015learning,gao2015you} that
use features from the last inner product layer, we choose the features $f_{I}$
from the last pooling layer, which retains spatial information of the original
images. We first rescale the images to be $448\times 448$ pixels, and then take
the features from the last pooling layer, which therefore have a dimension of
$512\times 14\times 14$, as shown in Fig.~\ref{fig:cnn_img}. $14\times 14$ is
the number of regions in the image and $512$ is the dimension of the feature
vector for each region. Accordingly, each feature vector in $f_{I}$ corresponds
to a $32\times 32$ pixel region of the input images. We denote by
$f_i, i\in[0, 195]$ the feature vector of each image region.

Then for modeling convenience, we use a single layer perceptron to transform
each feature vector to a new vector that has the same dimension as the question
vector (described in Sec.~\ref{sec:question_model}):
\begin{align}
  v_I = \tanh(W_If_I + b_I),
\end{align}
where $v_{I}$ is a matrix and its i-th column $v_{i}$ is the visual feature
vector for the region indexed by $i$.

\subsection{Question Model}
\label{sec:question_model}
As~\cite{sutskever2014sequence, shen2014latent, fang2014captions} show that
LSTMs and CNNs are powerful to capture the semantic meaning of texts, we
explore both models for question representations in this study.

\vspace{-0.2cm}
\subsubsection{LSTM based question model}
\label{sec:question_lstm}
\begin{figure}[tbh]
  \vspace{-0.6cm}
  \includegraphics[width=\linewidth]{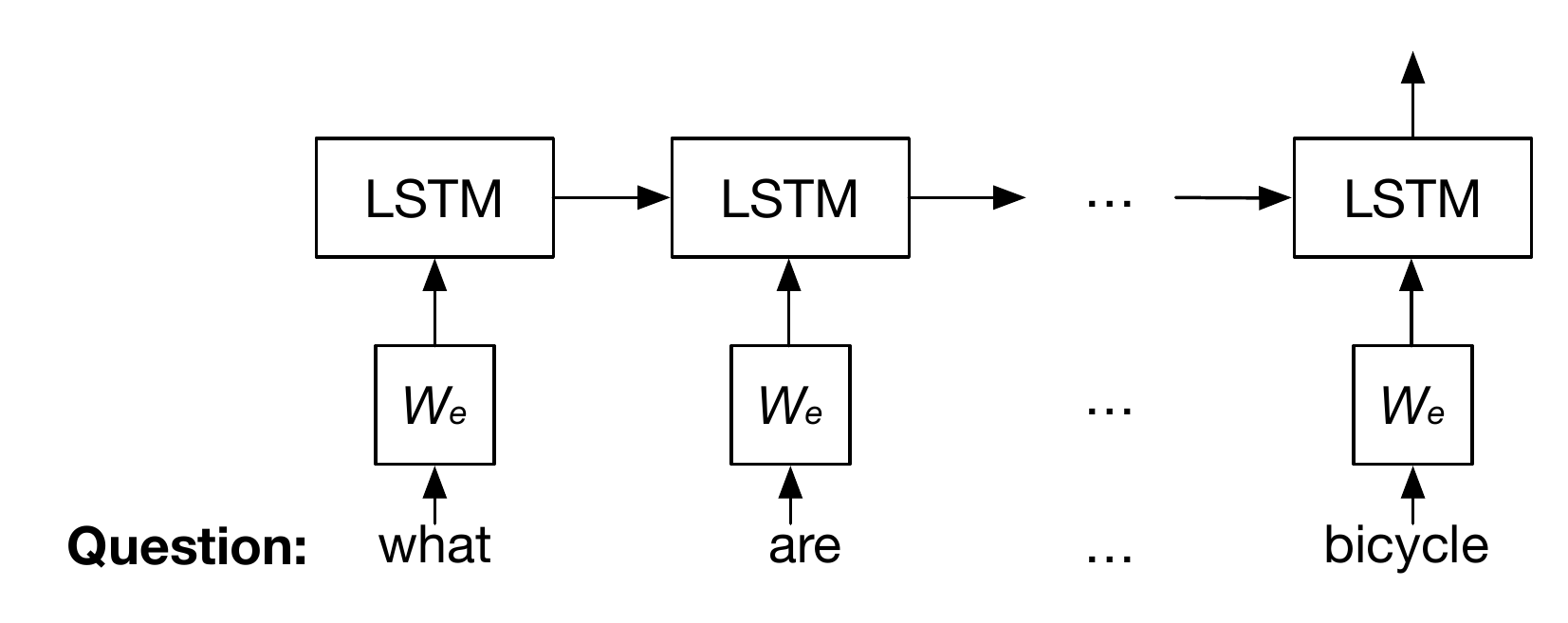}
  \centering
  \caption{LSTM based question model}
  \label{fig:lstm}
  \vspace{-0.2cm}
\end{figure}

The essential structure of a LSTM unit is a memory cell $c_{t}$ which reserves
the state of a sequence. At each step, the LSTM unit takes one input vector
(word vector in our case) $x_{t}$ and updates the memory cell $c_{t}$, then
output a hidden state $h_{t}$. The update process uses the gate mechanism. A
forget gate $f_{t}$ controls how much information from past state $c_{t-1}$ is
preserved. An input gate $i_{t}$ controls how much the current input $x_{t}$
updates the memory cell. An output gate $o_{t}$ controls how much information
of the memory is fed to the output as hidden state. The detailed update process
is as follows:
\begin{align}
  i_{t} =& \sigma(W_{xi}x_{t} + W_{hi}h_{t-1} + b_{i}), \\
  f_{t} =& \sigma(W_{xf}x_{t} + W_{hf}h_{t-1} + b_{f}), \\
  o_{t} =& \sigma(W_{xo}x_{t} + W_{ho}h_{t-1} + b_{o}), \\
  c_{t} =& f_{t}c_{t-1} + i_{t}\tanh(W_{xc}x_{t} + W_{hc}h_{t-1} + b_{c}), \\
  h_{t} =& o_{t}\tanh(c_{t}),
\end{align}
where $i, f, o, c$ are input gate, forget gate, output gate and memory cell,
respectively. The weight matrix and bias are parameters of the LSTM and are
learned on training data.

Given the question $q = [q_{1}, ...q_{T}]$, where $q_{t}$ is the one hot vector
representation of word at position $t$, we first embed the words to a vector
space through an embedding matrix $x_{t} = W_{e}q_{t}$. Then for every time
step, we feed the embedding vector of words in the question to LSTM:
\begin{align}
  x_{t} =& W_{e}q_{t}, t\in \{1,2,...T\}, \\
  h_{t} =& \text{LSTM}(x_{t}), t\in \{1,2,...T\}.
\end{align}

As shown in Fig.~\ref{fig:lstm}, the question \texttt{what are sitting in the
  basket on a bicycle} is fed into the LSTM. Then the final hidden layer is
taken as the representation vector for the question, i.e., $v_{Q} = h_{T}$.

\subsubsection{CNN based question model}
\label{sec:question_cnn}
\begin{figure}[tbh]
  \vspace{-0.3cm}
  \includegraphics[width=1.1\linewidth]{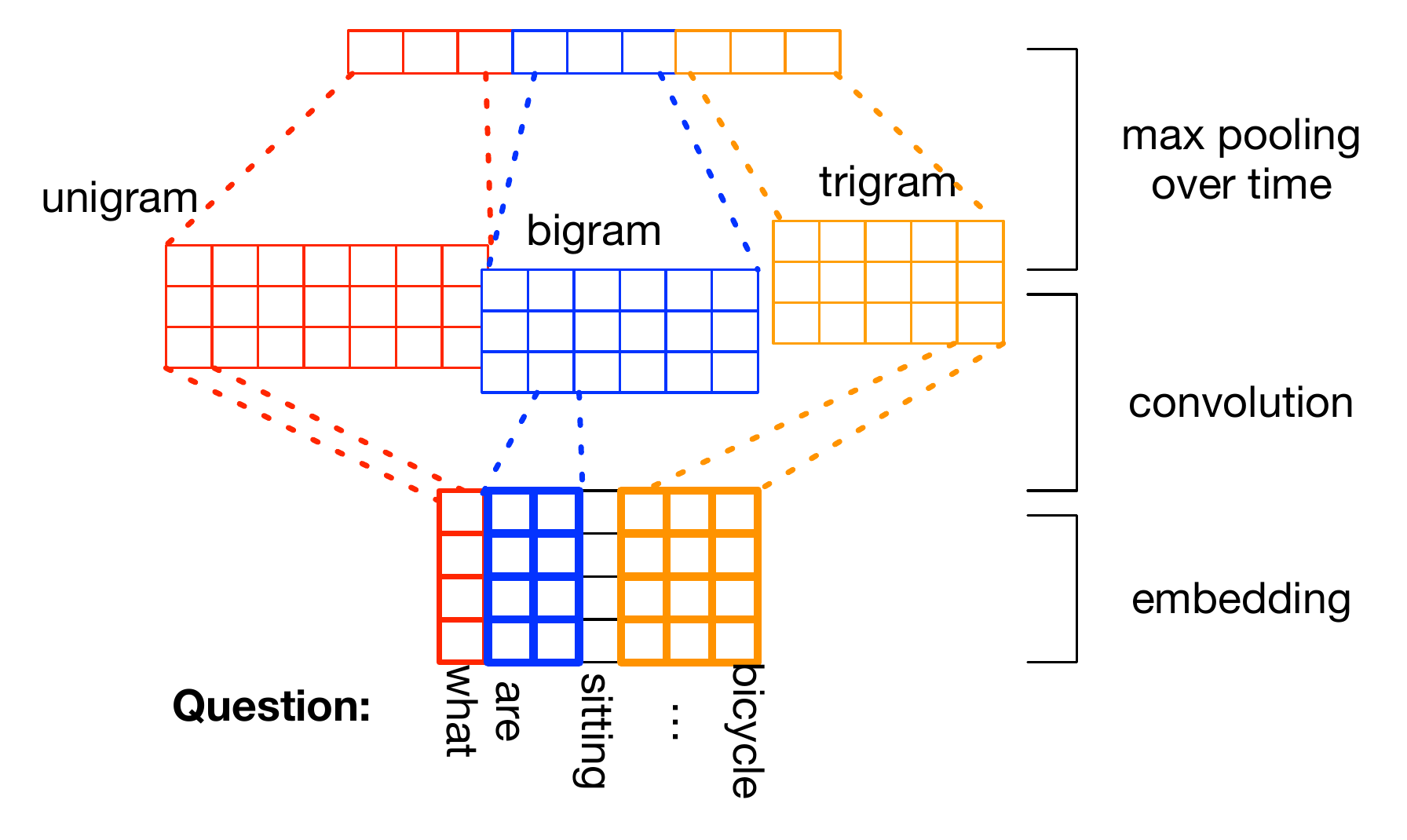}
  \centering
  \caption{CNN based question model}
  \label{fig:cnn}
  \vspace{-0.3cm}
\end{figure}

In this study, we also explore to use a CNN similar
to~\cite{kim2014convolutional} for question representation. Similar to the
LSTM-based question model, we first embed words to vectors $x_{t} = W_{e}q_{t}$
and get the question vector by concatenating the word vectors:
\begin{align}
  \vspace{-0.5cm}
  x_{1:T} = [x_{1}, x_{2}, ..., x_{T}].
  \vspace{-0.5cm}
\end{align}
Then we apply convolution operation on the word embedding vectors. We use three
convolution filters, which have the size of one (unigram), two (bigram) and
three (trigram) respectively. The $t$-th convolution output using window size
$c$ is given by:
\begin{align}
  h_{c,t} = \tanh(W_{c}x_{t:t+c-1} + b_{c}).
\end{align}
The filter is applied only to window $t:t+c-1$ of size $c$. $W_{c}$ is the
convolution weight and $b_{c}$ is the bias. The feature map of the filter with
convolution size $c$ is given by:
\begin{align}
  h_{c} = [h_{c,1}, h_{c,2}, ..., h_{c,T-c+1}].
\end{align}

Then we apply max-pooling over the feature maps of the convolution size $c$ and
denote it as
\begin{align}
  \tilde{h}_c = \max_{t}[h_{c,1}, h_{c,2}, ..., h_{c,T-c+1}].
\end{align}
The max-pooling over these vectors is a coordinate-wise max operation.  For
convolution feature maps of different sizes $c= 1, 2, 3$, we concatenate them
to form the feature representation vector of the whole question sentence:
\begin{align}
  h = [\tilde{h}_{1}, \tilde{h}_{2}, \tilde{h}_{3}],
\end{align}
hence $v_{Q} = h$ is the CNN based question vector.

The diagram of CNN model for question is shown in Fig.~\ref{fig:cnn}. The
convolutional and pooling layers for unigrams, bigrams and trigrams are drawn
in red, blue and orange, respectively.

\subsection{Stacked Attention Networks}
\label{sec:deep_attention_network}

Given the image feature matrix $v_{I}$ and the question feature vector $v_{Q}$,
SAN predicts the answer via multi-step reasoning.

In many cases, an answer only related to a small region of an image. For
example, in Fig.~\ref{fig:example}, although there are multiple objects in the
image: \texttt{bicycles, baskets, window, street} and \texttt{dogs} and the
answer to the question only relates to \texttt{dogs}. Therefore, using the one
global image feature vector to predict the answer could lead to sub-optimal
results due to the noises introduced from regions that are irrelevant to the
potential answer. Instead, reasoning via multiple attention layers
progressively, the SAN are able to gradually filter out noises and pinpoint the
regions that are highly relevant to the answer.

Given the image feature matrix $v_{I}$ and the question vector $v_{Q}$, we
first feed them through a single layer neural network and then a softmax
function to generate the attention distribution over the regions of the image:
\begin{align}
  h_{A} =& \tanh(W_{I, A}v_{I} \oplus (W_{Q,A}v_{Q}  + b_{A})), \\
  p_{I} = &\text{softmax}(W_Ph_{A} + b_{P}),
\end{align}
where $v_{I}\in \mathbf{R}^{d\times m}$, $d$ is the image representation
dimension and $m$ is the number of image regions, $v_{Q}\in \mathbf{R}^{d}$ is
a $d$ dimensional vector. Suppose $W_{I,A} ,W_{Q,A} \in \mathbf{R}^{k\times d}$
and $W_{P}\in \mathbf{R}^{1\times k}$, then $p_{I}\in \mathbf{R}^{m}$ is an $m$
dimensional vector, which corresponds to the attention probability of each
image region given $v_{Q}$. Note that we denote by $\oplus$ the addition of a
matrix and a vector. Since $W_{I, A}v_{I}\in \mathbf{R}^{k\times m}$ and both
$W_{Q, A}v_{Q}, b_{A} \in \mathbf{R}^{k}$ are vectors, the addition between a
matrix and a vector is performed by adding each column of the matrix by the
vector.

Based on the attention distribution, we calculate the weighted sum of the image
vectors, each from a region, $\tilde{v}_{i}$ as in
Eq.~\ref{eq:weighted_sum}. We then combine $\tilde{v}_{i}$ with the question
vector $v_{Q}$ to form a refined query vector $u$ as in Eq.~\ref{eq:query}. $u$
is regarded as a refined query since it encodes both question information and
the visual information that is relevant to the potential answer:
\begin{align}
  \vspace{-0.5cm}
  \tilde{v}_{I} =& \sum_{i}p_{i}v_{i}, \label{eq:weighted_sum} \\
  u =& \tilde{v}_{I} + v_{Q}. \label{eq:query}
\end{align}

Compared to models that simply combine the question vector and the global image
vector, attention models construct a more informative $u$ since higher weights
are put on the visual regions that are more relevant to the question. However,
for complicated questions, a single attention layer is not sufficient to locate
the correct region for answer prediction. For example, the question in
Fig.~\ref{fig:model_example} \texttt{what are sitting in the basket on a
  bicycle} refers to some subtle relationships among multiple objects in an
image. Therefore, we iterate the above query-attention process using multiple
attention layers, each extracting more fine-grained visual attention
information for answer prediction. Formally, the SANs take the following
formula: for the $k$-th attention layer, we compute:
\begin{align}
  h^{k}_{A} =& \tanh(W_{I, A}^{k}v_{I} \oplus (W_{Q,A}^{k}u^{k-1}  + b_{A}^{k})), \\
  p^k_{I} = &\text{softmax}(W_P^{k}h^{k}_{A} + b_{P}^{k}).
\end{align}
where $u^{0}$ is initialized to be $v_{Q}$. Then the aggregated image feature
vector is added to the previous query vector to form a new query vector:
\begin{align}
  \tilde{v}_{I}^{k} =&\sum_{i} p_{i}^{k} v_{i}, \\
  u^{k} =& \tilde{v}_{I}^{k} + u^{k-1}.
\end{align}

That is, in every layer, we use the combined question and image vector
$u^{k-1}$ as the query for the image. After the image region is picked, we
update the new query vector as $u^{k} = \tilde{v}_{I}^{k} + u^{k-1}$. We repeat
this $K$ times and then use the final $u^{K}$ to infer the answer:
\begin{align}
  \vspace{-0.2cm}
  p_{\text{ans}} =& \text{softmax}(W_{u}u^{K} + b_{u}).
\end{align}

Fig.~\ref{fig:example} illustrates the reasoning process by an example. In the
first attention layer, the model identifies roughly the area that are relevant
to \texttt{basket, bicycle}, and \texttt{sitting in}. In the second attention
layer, the model focuses more sharply on the region that corresponds to the
answer \texttt{dogs}. More examples can be found in Sec.~\ref{sec:experiments}.

\section{Experiments}
\label{sec:experiments}
\subsection{Data sets}
We evaluate the SAN on four image QA data sets.

{\bf DAQUAR-ALL} is proposed in \cite{malinowski2014multi}. There are $6,795$
training questions and $5,673$ test questions. These questions are generated on
$795$ and $654$ images respectively. The images are mainly indoor scenes. The
questions are categorized into three types including \emph{Object},
\emph{Color} and \emph{Number}. Most of the answers are single words. Following
the setting in \cite{ren2015imageqa, ma2015learning, malinowski2015ask}, we
exclude data samples that have multiple words answers. The remaining data set
covers $90\%$ of the original data set.

{\bf DAQUAR-REDUCED} is a reduced version of DAQUAR-ALL. There are $3,876$
training samples and $297$ test samples. This data set is constrained to $37$
object categories and uses only $25$ test images. The single word answers data
set covers $98\%$ of the original data set.

{\bf COCO-QA} is proposed in \cite{ren2015imageqa}. Based on the Microsoft COCO
data set, the authors first parse the caption of the image with an
off-the-shelf parser, then replace the key components in the caption with
question words for form questions. There are $78736$ training samples and
$38948$ test samples in the data set. These questions are based on $8,000$ and
$4,000$ images respectively. There are four types of questions including
\emph{Object}, \emph{Number}, \emph{Color}, and \emph{Location}. Each type
takes $70\%, 7\%, 17\%$, and $6\%$ of the whole data set, respectively. All
answers in this data set are single word.

{\bf VQA} is created through human labeling \cite{antol2015vqa}. The data set
uses images in the COCO image caption data set~\cite{lin2014microsoft}. Unlike
the other data sets, for each image, there are three questions and for each
question, there are ten answers labeled by human annotators. There are
$248,349$ training questions and $121,512$ validation questions in the data %
set. Following \cite{antol2015vqa}, we use the top $1000$ most frequent answer
as possible outputs and this set of answers covers $82.67\%$ of all answers.
We first studied the performance of the proposed model on the validation set.
Following \cite{fang2014captions}, we split the validation data set into two
halves, val1 and val2. We use training set and val1 to train and validate and
val2 to test locally. The results on the val2 set are reported in
Table.~\ref{tab:vqa}. We also evaluated the best model, SAN(2, CNN), on the
standard test server as provided in \cite{antol2015vqa} and report the results
in Table.~\ref{tab:vqa_server}.


\subsection{Baselines and evaluation methods}
We compare our models with a set of baselines proposed
recently~\cite{ren2015imageqa, antol2015vqa, malinowski2014multi,
  malinowski2015ask, ma2015learning} on image QA. Since the results of these
baselines are reported on different data sets in different literature, we
present the experimental results on different data sets in different tables.

For all four data sets, we formulate image QA as a classification problem since
most of answers are single words. We evaluate the model using classification
accuracy as reported in \cite{antol2015vqa, ren2015imageqa,
  malinowski2015ask}. The reference models also report the Wu-Palmer similarity
(WUPS) measure \cite{wu1994verbs}. The WUPS measure calculates the similarity
between two words based on their longest common subsequence in the taxonomy
tree. We can set a \mbox{threshold} for WUPS, if the similarity is less than
the threshold, then it is zeroed out. Following the reference models, we use
WUPS0.9 and WUPS0.0 as evaluation metrics besides the classification
accuracy. The evaluation on the VQA data set is different from other three data
sets, since for each question there are ten answer labels that may or may not
be the same. We follow~\cite{antol2015vqa} to use the following metric:
$min( \text{\# human labels that match that answer}/ 3 , 1)$, which basically
gives full credit to the answer when three or more of the ten human labels
match the answer and gives partial credit if there are less matches.

\subsection{Model configuration and training}
For the image model, we use the VGGNet to extract features. When training the
SAN, the parameter set of the CNN of the VGGNet is fixed. We take the output
from the last pooling layer as our image feature which has a dimension of
$512\times 14 \times 14$ .

For DAQUAR and COCO-QA, we set the word embedding dimension and LSTM's
dimension to be $500$ in the question model. For the CNN based question model,
we set the unigram, bigram and trigram convolution filter size to be $128$,
$256$, $256$ respectively. The combination of these filters makes the question
vector size to be $640$. For VQA dataset, since it is larger than other data
sets, we double the model size of the LSTM and the CNN to accommodate the large
data set and the large number of classes. In evaluation, we experiment with SAN
with one and two attention layers. We find that using three or more attention
layers does not further improve the performance.

In our experiments, all the models are trained using stochastic gradient
descent with momentum $0.9$. The batch size is fixed to be $100$. The best
learning rate is picked using grid search. Gradient clipping
technique~\cite{graves2013generating} and dropout~\cite{srivastava2014dropout}
are used.

\subsection{Results and analysis }
The experimental results on DAQUAR-ALL, DAQUAR-REDUCED, COCO-QA and VQA are
presented in Table.~\ref{tab:daquar_all_results} to ~\ref{tab:vqa}
respectively. Our model names explain their settings: SAN is short for the
proposed stacked attention networks, the value 1 or 2 in the brackets refer to
using one or two attention layers, respectively. The keyword LSTM or CNN refers
to the question model that SANs use.

\begin{table}[!thbp]
  \vspace{-0.5cm}
      \centering
  \begin{tabular}{l  c  c  c }
    \toprule
    Methods & Accuracy & WUPS0.9 &  WUPS0.0
    \\
    \midrule
    \multicolumn{4}{l}{
    {\bf \texttt{Multi-World:}} \cite{malinowski2014multi}} \\
    Multi-World & 7.9 & 11.9 & 38.8 \\
    \midrule
    \multicolumn{4}{l}{
    {\bf \texttt{Ask-Your-Neurons:}} \cite{malinowski2015ask}} \\
    Language & 19.1 & 25.2 & 65.1 \\
    Language + IMG & 21.7 & 28.0 & 65.0 \\
    {\bf \texttt{CNN:}} \cite{ma2015learning} \\
    IMG-CNN & 23.4 & 29.6 & 63.0 \\
    \midrule
    {\bf \texttt{Ours:}} \\
    SAN(1, LSTM) & 28.9  & 34.7 & 68.5 \\
    SAN(1, CNN) & 29.2 & 35.1 & 67.8 \\
    SAN(2, LSTM) & {\bf 29.3} & 34.9 & 68.1 \\
    SAN(2, CNN) & {\bf 29.3} & {\bf 35.1} & {\bf 68.6} \\
    \midrule
    {\bf \texttt{Human :}\cite{malinowski2014multi}} \\
    Human & 50.2 & 50.8 & 67.3 \\
    \bottomrule
  \end{tabular}
  \caption{DAQUAR-ALL results, in percentage}
  \label{tab:daquar_all_results}
  \end{table}
\begin{table}
      \centering
  \begin{tabular}{l  c  c  c }
    \toprule
    Methods & Accuracy & WUPS0.9 &  WUPS0.0
    \\
    \midrule
    \multicolumn{4}{l}{
    {\bf \texttt{Multi-World:}} \cite{malinowski2014multi}} \\
    Multi-World & 12.7 & 18.2 & 51.5 \\
    \midrule
    \multicolumn{4}{l}{
    {\bf \texttt{Ask-Your-Neurons:}} \cite{malinowski2015ask}} \\
    Language & 31.7 & 38.4 & 80.1 \\
    Language + IMG & 34.7 & 40.8 & 79.5 \\
    \midrule
    {\bf\texttt{VSE:}} \cite{ren2015imageqa} \\
    GUESS & 18.2 & 29.7 & 77.6 \\
    BOW & 32.7 & 43.2 & 81.3 \\
    LSTM & 32.7 & 43.5 & 81.6 \\
    IMG+BOW & 34.2 & 45.0 & 81.5 \\
    VIS+LSTM & 34.4 & 46.1 & 82.2 \\
    2-VIS+BLSTM & 35.8 & 46.8 & 82.2 \\
    \midrule
    {\bf \texttt{CNN:}} \cite{ma2015learning} \\
    IMG-CNN & 39.7 & 44.9 & 83.1 \\
    \midrule
    {\bf \texttt{Ours:}} \\
    SAN(1, LSTM) & 45.2 & 49.6 & 84.0 \\
    SAN(1, CNN) & 45.2 & 49.6 & 83.7 \\
    SAN(2, LSTM) & {\bf 46.2} & {\bf 51.2} & {\bf 85.1} \\
    SAN(2, CNN) & 45.5 & 50.2 & 83.6 \\
    \midrule
    {\bf \texttt{Human :}\cite{malinowski2014multi}} \\
    Human & 60.3 & 61.0 & 79.0 \\
    \bottomrule
  \end{tabular}
  \caption{DAQUAR-REDUCED results, in percentage}
  \label{tab:daquar_reduced_results}
  \vspace{-0.5cm}
\end{table}

\begin{table}[!thbp]
  \vspace{-0.5cm}
      \centering
  \begin{tabular}{l  c  c  c }
    \toprule
    Methods & Accuracy & WUPS0.9 &  WUPS0.0
    \\
    \midrule
    {\bf\texttt{VSE:}} \cite{ren2015imageqa} \\
    GUESS  & 6.7 & 17.4 & 73.4 \\
    BOW & 37.5 & 48.5 & 82.8 \\
    LSTM & 36.8 & 47.6 & 82.3 \\
    IMG & 43.0 & 58.6 & 85.9 \\
    IMG+BOW & 55.9 & 66.8 & 89.0 \\
    VIS+LSTM & 53.3 & 63.9 & 88.3 \\
    2-VIS+BLSTM & 55.1 & 65.3 & 88.6 \\
    \midrule
    {\bf \texttt{CNN:}} \cite{ma2015learning} \\
    IMG-CNN & 55.0 & 65.4 & 88.6 \\
    CNN & 32.7 & 44.3 & 80.9 \\
    \midrule
    {\bf \texttt{Ours:}} \\
    SAN(1, LSTM) & 59.6 & 69.6 & 90.1 \\
    SAN(1, CNN) & 60.7 & 70.6 & 90.5 \\
    SAN(2, LSTM) & 61.0 & 71.0 & 90.7 \\
    SAN(2, CNN) & {\bf 61.6} & {\bf 71.6} & {\bf 90.9} \\
    \bottomrule
  \end{tabular}
\caption{COCO-QA results, in percentage}
\label{tab:coco_results}
\end{table}
\begin{table}
      \centering
  \begin{tabular}{l  c  c  c  c}
    \toprule
    Methods & Objects & Number & Color & Location \\
    \midrule
    {\bf\texttt{VSE:}} \cite{ren2015imageqa} \\
    GUESS & 2.1 & 35.8 & 13.9 & 8.9 \\
    BOW & 37.3 & 43.6 & 34.8 & 40.8 \\
    LSTM & 35.9 & 45.3 & 36.3 & 38.4 \\
    IMG & 40.4 & 29.3 & 42.7 & 44.2 \\
    IMG+BOW & 58.7 & 44.1 & 52.0 & 49.4 \\
    VIS+LSTM & 56.5 & 46.1 & 45.9 & 45.5 \\
    2-VIS+BLSTM & 58.2 & 44.8 & 49.5 & 47.3 \\
    \midrule
    {\bf \texttt{Ours:}} \\
    SAN(1, LSTM) & 62.5 & 49.0 & 54.8 & 51.6 \\
    SAN(1, CNN) & 63.6 & 48.7 & 56.7 & 52.7 \\
    SAN(2, LSTM) & 63.6 & {\bf 49.8} & 57.9 & 52.8 \\
    SAN(2, CNN) & {\bf 64.5} & 48.6 & {\bf 57.9} & {\bf 54.0} \\
    \bottomrule
  \end{tabular}
  \caption{COCO-QA accuracy per class, in percentage}
  \label{tab:coco_perclass}
  \vspace{-0.5cm}
\end{table}

\begin{table}
  \centering
  \small
  \begin{tabular}{l c c c c c}
    \toprule
    & \multicolumn{4}{c}{test-dev} & test-std \\
    \cline{2-5}\\
    Methods & All & Yes/No & Number & Other & All\\
    \midrule
    {\bf\texttt{VQA:}} \cite{antol2015vqa} \\
    Question & 48.1 & 75.7 & 36.7 & 27.1 & - \\
    Image & 28.1 & 64.0 & 0.4 & 3.8 & -\\
    Q+I & 52.6 & 75.6 & 33.7 & 37.4 & -\\
    LSTM Q & 48.8 & 78.2 & 35.7 & 26.6 & -\\
    LSTM Q+I & 53.7 & 78.9 & 35.2 & 36.4 & 54.1 \\
    \midrule
    SAN(2, CNN) & {\bf 58.7} & {\bf 79.3} & {\bf 36.6} & {\bf 46.1} & {\bf 58.9}\\
    \bottomrule
  \end{tabular}
  \caption{VQA results on the official server, in percentage}
\label{tab:vqa_server}
\vspace{-0.6cm}
\end{table}

\begin{table}
  \centering
  \begin{tabular}{l  c c c c}
    \toprule
    Methods & All & \shortstack{Yes/No\\ 36\%} & \shortstack{Number\\ 10\%} &
                                                                              \shortstack{Other\\ 54\%}\\
    \midrule
    SAN(1, LSTM) & 56.6 & 78.1 & 41.6 & 44.8\\
    SAN(1, CNN) & 56.9 & 78.8 & 42.0 & 45.0\\
    SAN(2, LSTM) & 57.3 & 78.3 & {\bf 42.2} & 45.9\\
    SAN(2, CNN)& {\bf 57.6} & 78.6 & 41.8 & {\bf 46.4}\\
    \bottomrule
  \end{tabular}
  \caption{VQA results on our partition, in percentage}
\label{tab:vqa}
\vspace{-0.6cm}
\end{table}


The experimental results in Table.~\ref{tab:daquar_all_results} to
\ref{tab:vqa} show that the two-layer SAN gives the best results across all
data sets and the two kinds of question models in the SAN, LSTM and CNN, give
similar performance. For example, on DAQUAR-ALL
(Table.~\ref{tab:daquar_all_results}), both of the proposed two-layer SANs
outperform the two best baselines, the IMG-CNN in \cite{ma2015learning} and the
Ask-Your-Neuron in \cite{malinowski2015ask}, by $5.9\%$ and $7.6\%$ absolute in
accuracy, respectively. Similar range of improvements are observed in metrics
of WUPS0.9 and WUPS0.0. We also observe significant improvements on
DAQUAR-REDUCED (Table.~\ref{tab:daquar_reduced_results}), i.e., our SAN(2,
LSTM) outperforms the IMG-CNN \cite{ma2015learning}, the 2-VIS+BLSTM
\cite{ren2015imageqa}, the Ask-Your-Neurons approach \cite{malinowski2015ask}
and the Multi-World \cite{malinowski2014multi} by $6.5\%$, $10.4\%$, $11.5\%$
and $33.5\%$ absolute in accuracy, respectively. On the larger COCO-QA data
set, the proposed two-layer SANs significantly outperform the best baselines
from \cite{ma2015learning} (IMG-CNN) and \cite{ren2015imageqa} (IMG+BOW and
2-VIS+BLSTM) by 5.1\% and 6.6\% in accuracy
(Table.~\ref{tab:coco_results}). Table.~\ref{tab:vqa_server} summarizes the
performance of various models on VQA, which is the largest among the four data
sets. The overall results show that our best model, SAN(2, CNN), outperforms
the LSTM Q+I model, the best baseline from \cite{antol2015vqa}, by 4.8\%
absolute. The superior performance of the SANs across all four benchmarks
demonstrate the effectiveness of using multiple layers of attention.

In order to study the strength and weakness of the SAN in detail, we report
performance at the question-type level on the two large data sets, COCO-QA and
VQA, in Table.~\ref{tab:coco_perclass} and ~\ref{tab:vqa_server},
respectively. We observe that on COCO-QA, compared to the two best baselines,
IMG+BOW and 2-VIS+BLSTM, out best model SAN(2, CNN) improves 7.2\% in the
question type of \emph{Color}, followed by 6.1\% in \emph{Objects}, 5.7\% in
\emph{Location} and 4.2\% in \emph{Number}. We observe similar trend of
improvements on VQA. As shown in Table.~\ref{tab:vqa_server}, compared to the
best baseline LSTM Q+I, the biggest improvement of SAN(2, CNN) is in the
\emph{Other} type, 9.7\%, followed by the 1.4\% improvement in \emph{Number}
and 0.4\% improvement in \emph{Yes/No}. Note that the \emph{Other} type in VQA
refers to questions that usually have the form of ``what color, what kind, what
are, what type, where'' etc., which are similar to question types of
\emph{Color}, \emph{Objects} and \emph{Location} in COCO-QA. The VQA data set
has a special \emph{Yes/No} type of questions. The SAN only improves the
performance of this type of questions slightly. This could due to that the
answer for a \emph{Yes/No} question is very question dependent, so better
modeling of the visual information does not provide much additional gains. This
also confirms the similar observation reported in \cite{antol2015vqa}, e.g.,
using additional image information only slightly improves the performance in
\emph{Yes/No}, as shown in Table.~\ref{tab:vqa_server}, Q+I vs Question, and
LSTM Q+I vs LSTM Q.

Our results demonstrate clearly the positive impact of using multiple attention
layers. In all four data sets, two-layer SANs always perform better than the
one-layer SAN. Specifically, on COCO-QA, on average the two-layer SANs
outperform the one-layer SANs by 2.2\% in the type of \emph{Color}, followed by
1.3\% and 1.0\% in the \emph{Location} and \emph{Objects} categories, and then
0.4\% in \emph{Number}. This aligns to the order of the improvements of the SAN
over baselines. Similar trends are observed on VQA (Table.~\ref{tab:vqa}),
e.g., the two-layer SAN improve over the one-layer SAN by 1.4\% for the
\emph{Other} type of question, followed by 0.2\% improvement for \emph{Number},
and flat for \emph{Yes/No}.

\subsection{Visualization of attention layers}
In this section, we present analysis to demonstrate that using multiple
attention layers to perform multi-step reasoning leads to more fine-grained
attention layer-by-layer in locating the regions that are relevant to the
potential answers. We do so by visualizing the outputs of the attention layers
of a sample set of images from the COCO-QA test set. Note the attention
probability distribution is of size $14\times 14$ and the original image is
$448\times448$, we up-sample the attention probability distribution and apply a
Gaussian filter to make it the same size as the original image.

Fig.~\ref{fig:vqa_more_examples} presents six examples. More examples are
presented in the appendix. They cover types as broad as \emph{Object},
\emph{Numbers}, \emph{Color} and \emph{Location}. For each example, the three
images from left to right are the original image, the output of the first
attention layer and the output of the second attention layer, respectively. The
bright part of the image is the detected attention. Across all those examples,
we see that in the first attention layer, the attention is scattered on many
objects in the image, largely corresponds to the objects and concepts referred
in the question, whereas in the second layer, the attention is far more focused
on the regions that lead to the correct answer. For example, consider the
question \texttt{what is the color of the horns}, which asks the color of the
horn on the woman's head in Fig.~\ref{fig:vqa_more_examples}(f). In the output
of the first attention layer, the model first recognizes a woman in the
image. In the output of the second attention layer, the attention is focused on
the head of the woman, which leads to the answer of the question: the color of
the horn is \texttt{red}.

\begin{figure*}[!tbhp]
  \vspace{-0.5cm}
  \centering
  \begin{minipage}{\textwidth}
      \centering
      \includegraphics[width=0.95\linewidth] {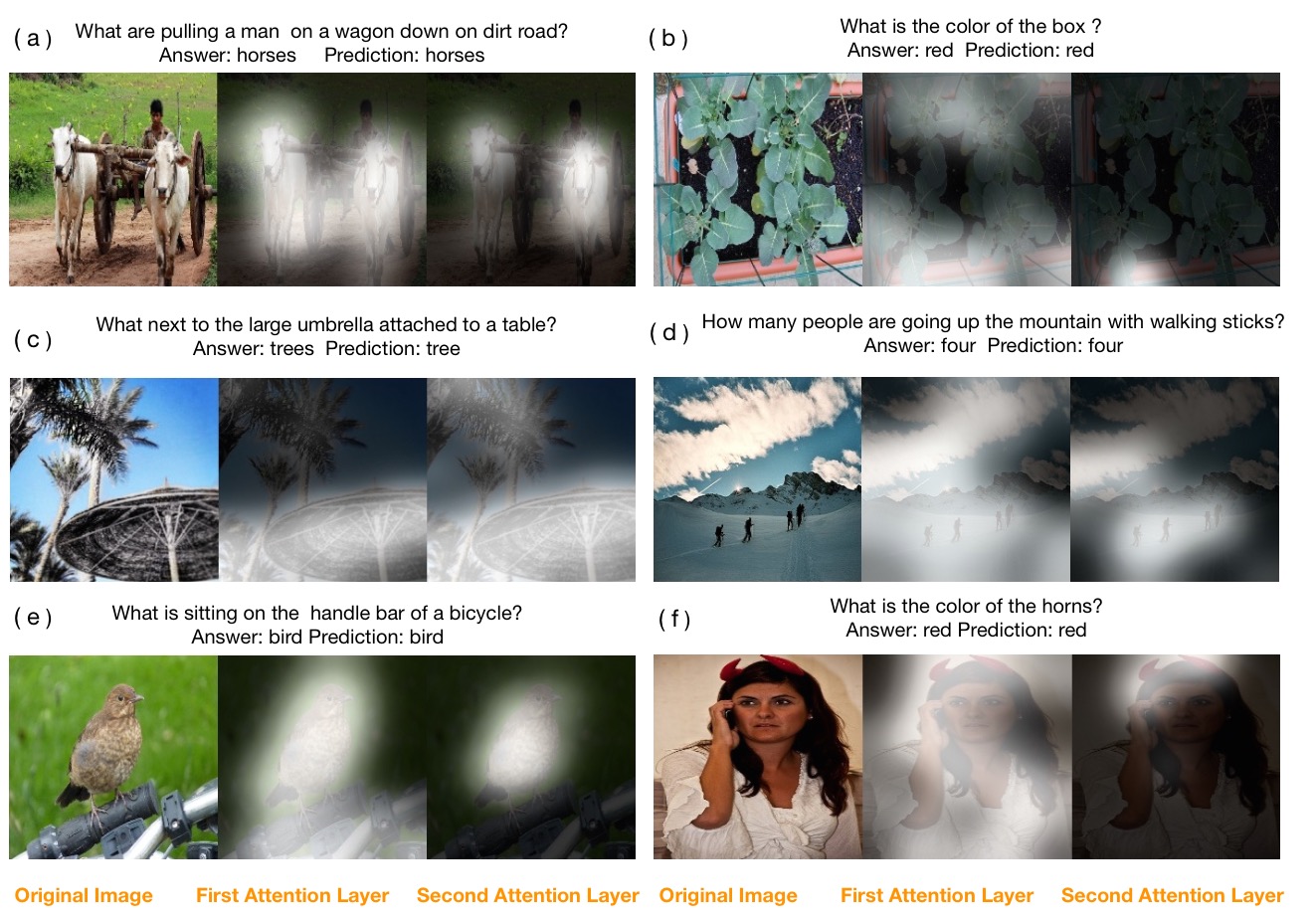}
      \centering
      \caption{Visualization of two attention layers}
      \label{fig:vqa_more_examples}
  \end{minipage}
  \centering
  \begin{minipage}{\textwidth}
    \includegraphics[width=0.95\linewidth] {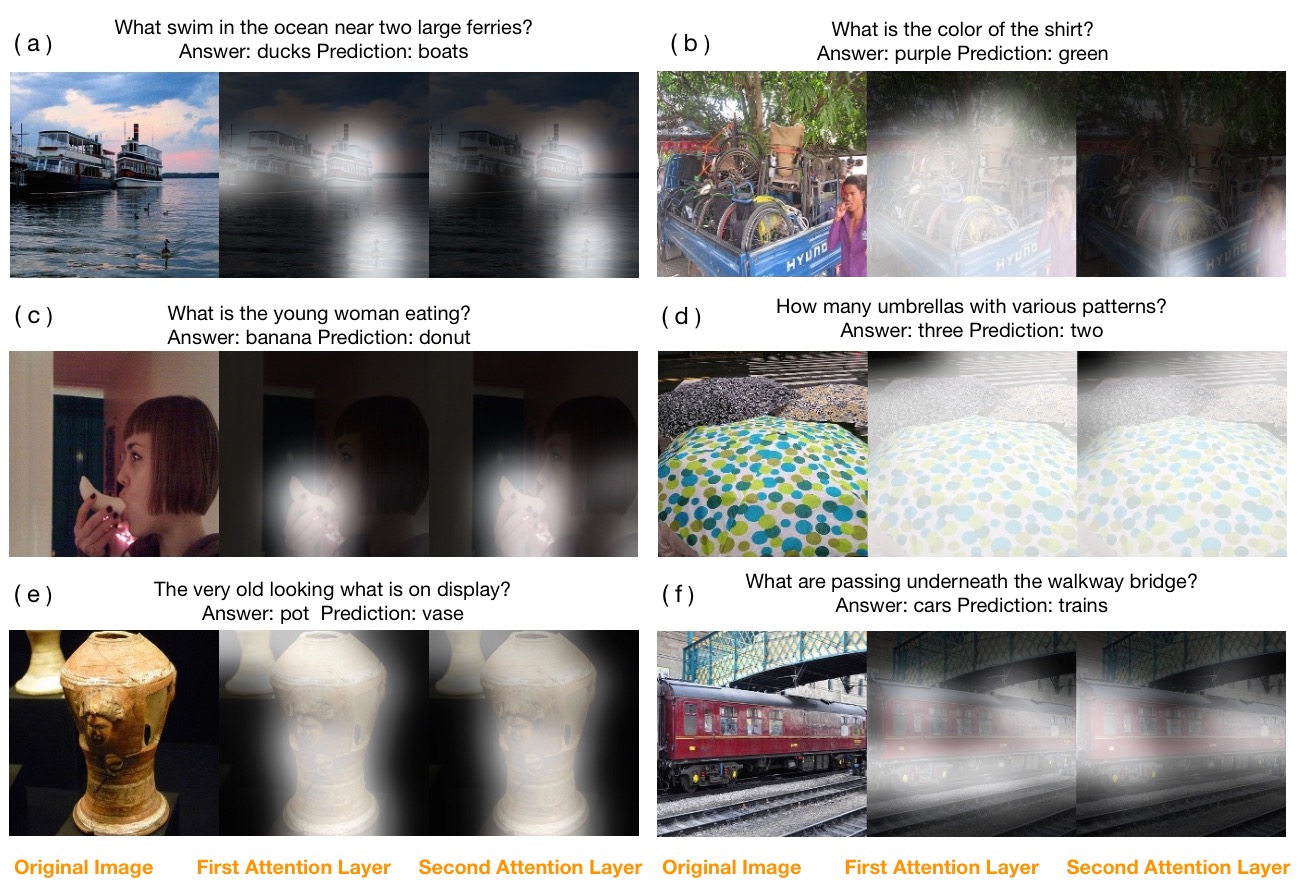}
    \centering
    \caption{Examples of mistakes}
    \label{fig:vqa_wrong_examples}
  \end{minipage}
\end{figure*}

\subsection{Errors analysis}
We randomly sample 100 images from the COCO-QA test set that the SAN make
mistakes. We group the errors into four categories: (i) the SANs focus the
attention on the wrong regions (22\%), e.g., the example in
Fig.~\ref{fig:vqa_wrong_examples}(a); (ii) the SANs focus on the right region
but predict a wrong answer (42\%), e.g., the examples in
Fig.~\ref{fig:vqa_wrong_examples}(b)(c)(d); (iii) the answer is ambiguous, the
SANs give answers that are different from labels, but might be acceptable
(31\%). E.g., in Fig.~\ref{fig:vqa_wrong_examples}(e), the answer label is
\texttt{pot}, but out model predicts \texttt{vase}, which is also visually
reasonable; (iv) the labels are clearly wrong (5\%). E.g., in
Fig.~\ref{fig:vqa_wrong_examples}(f), our model gives the correct answer
\texttt{trains} while the label \texttt{cars} is wrong.

\vspace{-0.2cm}
\section{Conclusion}
In this paper, we propose a new stacked attention network (SAN) for image
QA. SAN uses a multiple-layer attention mechanism that queries an image
multiple times to locate the relevant visual region and to infer the answer
progressively. Experimental results demonstrate that the proposed SAN
significantly outperforms previous state-of-the-art approaches by a substantial
margin on all four image QA data sets. The visualization of the attention
layers further illustrates the process that the SAN focuses the attention to
the relevant visual clues that lead to the answer of the question
layer-by-layer.


\newpage
{\small
\bibliographystyle{ieee}
\bibliography{bible}
}

\appendix
\begin{figure*}[!tbhp]
  \includegraphics[width=1.0\linewidth]
{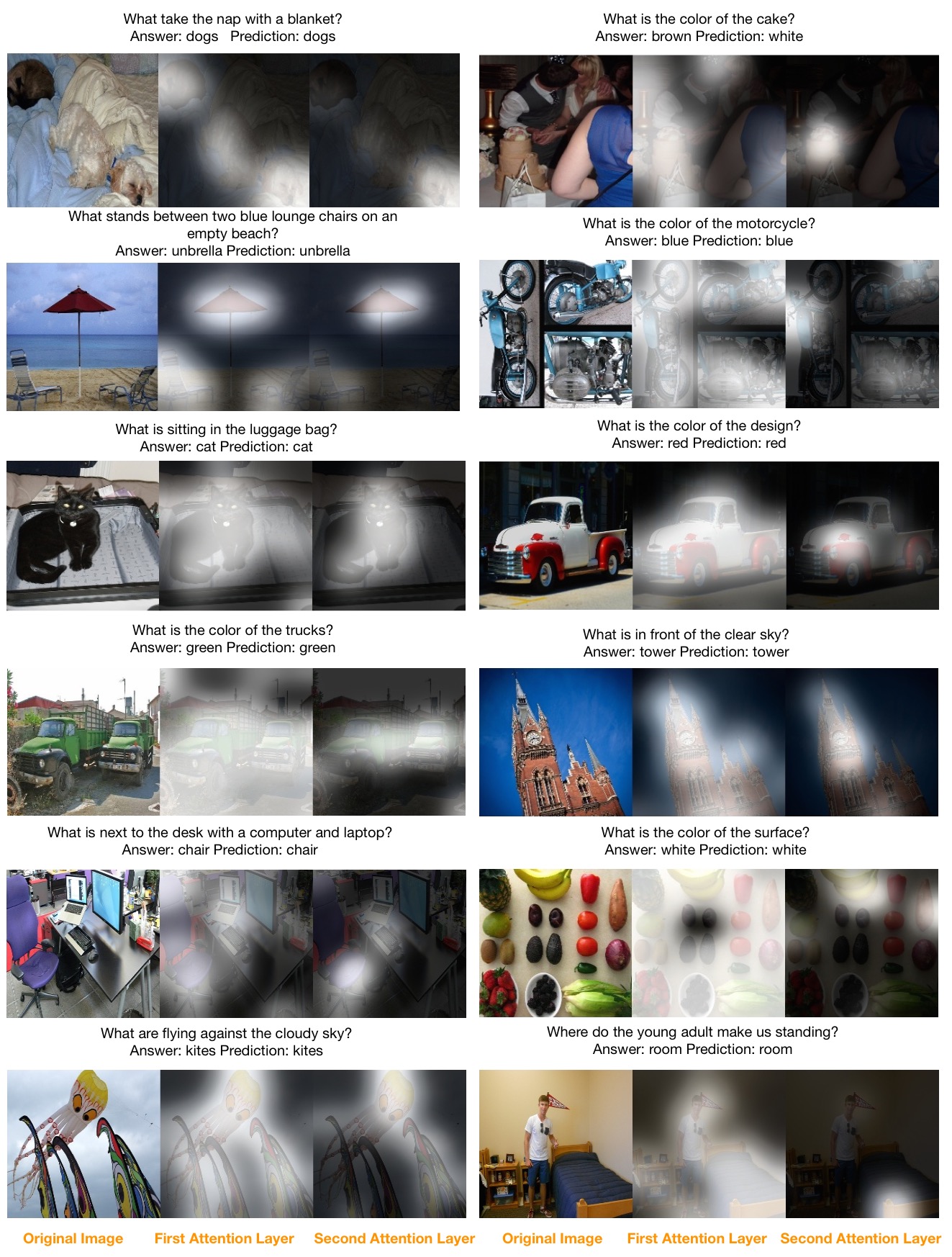}
  \centering
  \caption{More examples}
  \label{fig:vqa_examples_appendix}
\end{figure*}

\end{document}